\documentclass{article}

    \PassOptionsToPackage{numbers, compress,sort}{natbib}

\usepackage[preprint]{neurips_2021}

\usepackage{graphicx}

\usepackage[utf8]{inputenc} 
\usepackage[T1]{fontenc}    
\usepackage{hyperref}       
\usepackage{url}            
\usepackage{booktabs}       
\usepackage{amsfonts}       
\usepackage{nicefrac}       
\usepackage{microtype}      
\usepackage{xcolor}         

\usepackage{enumitem}
\setlist[enumerate,itemize]{itemsep=0.5pt,topsep=0pt}

\title{Thinking Fast and Slow in AI: \\
the Role of Metacognition}
\author{M. Bergamaschi Ganapini \\ Union College \And M. Campbell \\ IBM Research \And F. Fabiano \\ Univ. Udine\And L. Horesh  \\ IBM Research \And J. Lenchner \\ IBM Research \And A. Loreggia \\ EUI \And N. Mattei \\ Tulane Univ.\And F. Rossi \\ IBM Research \And B. Srivastava \\ USC\And K. B. Venable \\ IHMC}

\begin{document}

\maketitle



\vspace{-1cm}

\section{Overview} 

AI systems have seen dramatic advancement in recent years, bringing many applications that  pervade our everyday life. However, we are still mostly seeing instances of narrow AI: 
many of these recent developments are typically focused on a very limited set of competencies and goals, e.g., image interpretation, natural language processing, classification, prediction, and many others. Moreover, while these successes can be accredited to improved algorithms and techniques, they are also tightly linked to the availability of huge datasets and computational power \cite{marcus2020next}. State-of-the-art AI still lacks many capabilities that would naturally be included in a notion of (human) intelligence. Examples of these capabilities are generalizability, adaptability, robustness, explainability, causal analysis, abstraction, common sense reasoning, ethical reasoning \cite{RoMa19a}, as well as a complex and seamless integration of learning and reasoning supported by both implicit and explicit knowledge \cite{ai1002021}.

We argue that a better study of the mechanisms that allow humans to have these capabilities can help us understand how to imbue AI systems with these competencies \cite{rossi2019preferences,aaai2021-blue}. 
We focus especially on D. Kahneman's theory of thinking fast and slow  \cite{kahneman2011thinking}, and we propose a multi-agent AI architecture (called SOFAI, for SlOw and Fast AI) where incoming problems are solved by either system 1 (or "fast") agents (also called "solvers"), that react by exploiting only past experience, or by system 2 (or "slow") agents, 
that are deliberately activated when there
is the need to reason and search for optimal solutions beyond what is expected from the system 1 agent. 
Both kinds of agents are supported by a model of the world, containing domain knowledge about the environment, and a model of ``self'', containing information about past actions of the system and solvers' skills. 
Given the need to choose between these two kinds of solvers, a meta-cognitive agent is employed, performing introspection and arbitration roles, and assessing the need to employ system 2 solvers by considering resource constraints, abilities of the solvers, past experience, and expected reward for a correct solution of the given problem  \cite{Shenhav2013,Thompson2011}.
To do this balancing in a resource-conscious way, the meta-cognitive agent includes two successive phases, the first one faster and more approximate, and the second one (if needed) more careful and deliberate. 
Different approaches to the design of AI systems inspired by the dual-system theory have also been published recently
\cite{bengio2017consciousness,goel2017thinking,chen2019deep,anthony2017thinking,mittal2017thinking,DBLP:journals/ibmrd/NoothigattuBMCM19,gulati2020interleaving}.

Many real-life settings present sequential decision problems.
Depending on the availability of system 1 and/or system 2 solvers that can tackle single decisions or a sequence of them, the SOFAI architecture employs the meta-cognitive agent at each decision, or only once for a whole sequence \cite{DBLP:journals/ibmrd/NoothigattuBMCM19,glazier2021making,balakrishnan2018incorporating}. The first modality 
provides additional flexibility, since each call of the meta-cognitive module may choose a different solver to make the next decision, while  
the second one allows to exploit additional domain knowledge in the solvers. 

We hope that the SOFAI architecture will support more flexibility, faster performance, and higher decision quality than a single-modality system where meta-cognition is not employed, and/or where there is no distinction between system 1 and system 2 agents.
In this short paper, we describe the overall architecture and the role of the meta-cognitive agent. We provide motivation for the adopted design choices and we describe ongoing work to test instances of the architecture on sequential decision problems such as (epistemic) planning and path finding in constrained environments, against criteria intended to measure the quality of the decisions.

\section{Thinking Fast and Slow in Humans} 

According to Kahneman's theory, described in his book ``Thinking, Fast and Slow" \cite{kahneman2011thinking},
human's decisions are supported and guided by the cooperation of two kinds of capabilities, that, for sake of simplicity are called \emph{systems}: system 1 provides tools for intuitive, imprecise, fast, and often unconscious decisions (``thinking fast"), while system 2 handles more complex situations where logical and rational thinking is needed to reach a complex decision
(``thinking slow").

System 1 is guided mainly by intuition rather than deliberation. It gives fast answers to simple questions. Such answers are sometimes wrong, mainly because of unconscious bias or because they rely on heuristics or other short cuts \cite{gigerenzer2009}, and usually do not have explanations. 
However, system 1 is able to build models of the world that, although inaccurate and imprecise, can fill knowledge gaps through causal inference, allowing us to respond reasonably to the many stimuli of our everyday life.

When the problem is too complex for system 1, system 2 kicks in and solves it with access to additional computational resources, full attention, and sophisticated logical reasoning. A typical example of a problem handled by system 2 is solving a complex arithmetic calculation, or a multi-criteria optimization problem.
To do this, humans need to be able to recognize that a problem goes beyond a threshold of cognitive ease and therefore see the need to activate a more global and accurate reasoning machinery \cite{kahneman2011thinking}. Hence, introspection and meta-cognition is essential in this process.

When a problem is new and difficult to solve, it is handled by system 2 \cite{Kim2019}. However, certain problems over time pass on to system 1. The reason is that the procedures used by system 2 to find solutions to such problems are used to accumulate experience that system 1 can later use with little effort. Thus, over time, some problems, initially solvable only by resorting to system 2 reasoning tools, can become manageable by system 1. A typical example is reading text in our own native language. However, this does not happen with all tasks. An example of a problem that never passes to system 1 is finding the correct solution to complex arithmetic questions.

\section{Thinking Fast and Slow in AI: A Multi-agent Approach} 

\begin{figure}
  \centering
  \includegraphics[scale=0.28]{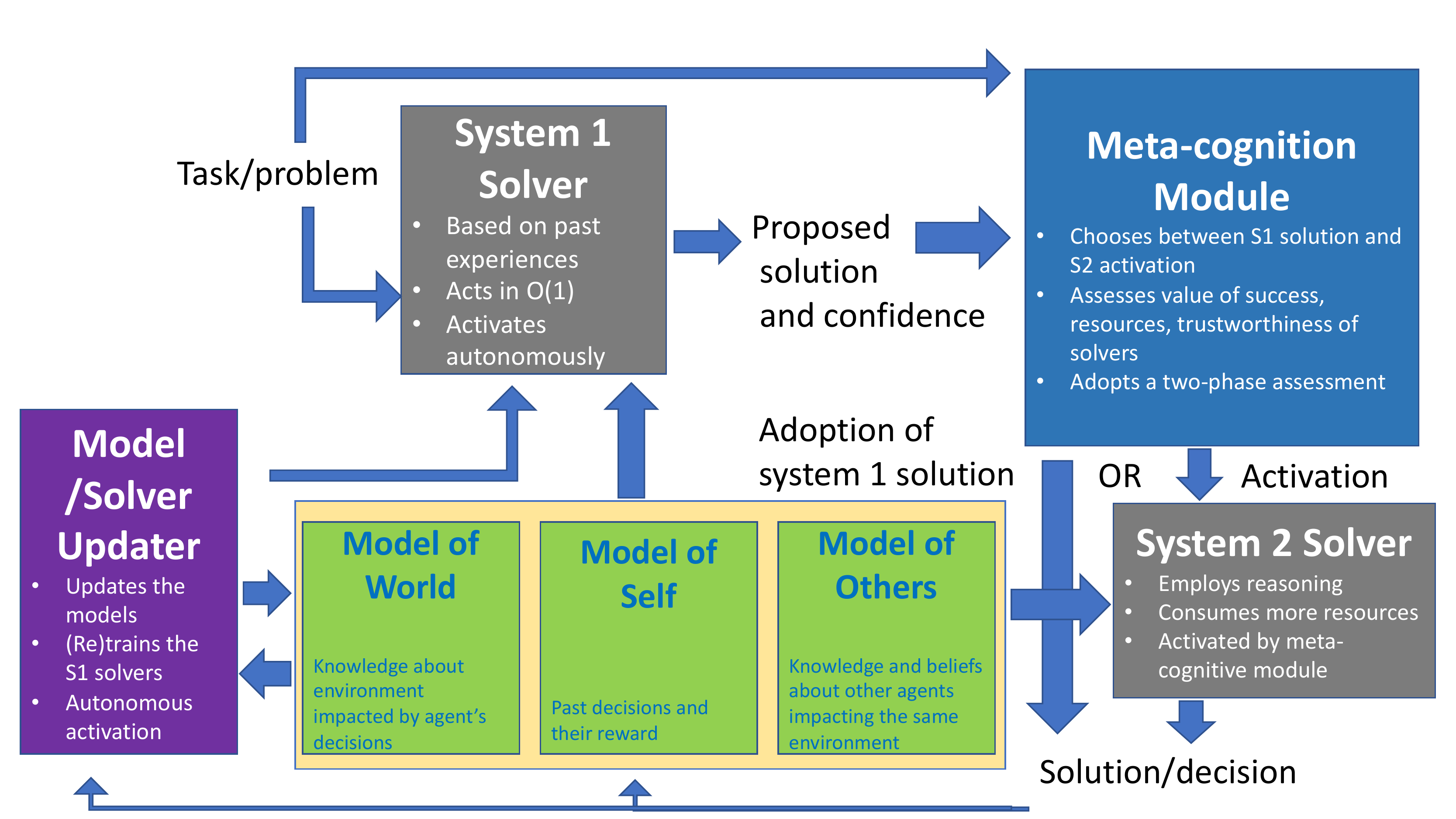}
  \caption{The SOFAI architecture, supporting system 1 / system 2 agents and meta-cognition.}
  \label{fig1}
\end{figure}

As shown in Figure \ref{fig1}, we are working on a multi-agent architecture which is inspired by the "Thinking Fast and Slow" theory of human decisions. In the SOFAI architectures, incoming problems/tasks are initially handled by system 1 (S1) solvers that have the required skills to tackle them, analogous to what is done by humans who unconsciously react to an external stimulus via their system 1 \cite{kahneman2011thinking}.

We assume S1 solvers act in constant time (that is, their running time is not a function of the size of the input problem) by relying on the past experience of the system, which is maintained in the model of self. 
The model of the world maintains the knowledge accumulated by the system over the external environment and the expected tasks, while the model of others maintains knowledge and beliefs over other agents that may act in the same environment. The model updater agent acts in the background to keep all models updated as new knowledge of the world, of other agents, or new decisions are generated and evaluated.

Once an S1 solver has solved the problem (for sake of simplicity, let's assume it is just one S1 solver), the proposed solution and the associated confidence level are available to the meta-cognitive (MC) module. At this point the MC agent starts its operations, with the task of choosing between adopting the S1 solver's solution or activating a system 2 (S2) solver. 
S2 agents use some form of reasoning over the current problem and usually consume more resources (especially time) than S1 agents. Also, they never work on a problem unless they are explicitly invoked by the MC module. 

To make its decision, the MC agent assesses the current resource availability, the expected resource consumption of the S2 solver, the expected reward for a correct solution and for each of the available solvers, as well as the solution and confidence evaluations coming from the S1 solver.
In order to not waste resources at the meta-cognitive level, the MC agent includes two successive assessment phases, the first one faster and more approximate, related to rapid unconscious assessment in humans \cite{Ackerman2017,Proust2004}, and the second one (to be used only if needed) more careful and resource-costly, analogous to the conscious introspective process in humans \cite{Carruthers}. 
The next section will provide more details about the internal steps of the MC agent. 



This is clearly an S1-by-default architecture, analogous to what happens in humans: whenever a new problem is presented, an S1 solver with the necessary skills to solve the problem starts working on it, generating a solution and a confidence level.
The MC agent will then decide if there is the need to activate an S2 solver. This allows for minimizing time to action (since S1 solvers act in constant time) when there is no need for S2 processing. It also
allows the MC agent to exploit the proposed action and confidence of S1
when deciding whether to activate S2, which leads to more informed and hopefully better decisions by MC.

Notice that 
we do not assume that S2 solvers are always better than S1 solvers,
analogously to what happens in humans \cite{gigerenzer2009}.
Take for example complex arithmetic, which usually requires humans to employ system 2, vs 
perception tasks, which are typically handled by our system 1. Similarly, in the SOFAI architecture we allow for tasks that are better handled by S1 solvers, especially once the system has acquired enough experience on those tasks.

\section{The Role of Meta-cognition} 

Meta-cognition is generally understood as any cognitive process that is about some other cognitive process \cite{CoxRaja2011}. We focus on the concept of meta-cognition as defined in \cite{Flavell1979, NN1990}, that is, the set of processes and mechanisms that could allow a computational system to both monitor and control its own cognitive activities, processes, and structures. The goal of this form of control is to improve the quality of the system’s decisions \cite{CoxRaja2011}. 
Among the existing computational models of meta-cognition \cite{Cox2005,Kralik2018,Posner2020},
we propose a centralized meta-cognitive module that exploits both internal and external data, and arbitrates between S1 vs S2 solvers in the process of solving a single task. Notice however that this arbitration is different than algorithm portfolio selection, 
which is successfully used for many problems \cite{kerschke2019automated},
because of the characterization of S1 and S2 solvers and the way the MC agent controls them.

The MC module 
exploits information coming from two main sources: 
\begin{enumerate}[wide, labelwidth=!, labelindent=5pt]
    \item The system’s internal models, which are periodically updated by the model updater agent:
    
    \begin{itemize}
        \item Model of self, which includes information about:
        \begin{itemize}
            \item Solvers’ past decisions in specific tasks.
            \item Resource consumption, e.g., memory and time of using solvers for specific tasks.
            \item Rewards of solvers' decisions. 
            \item Available system’s resources, e.g.,  memory, time.
            \item Expected reward for solving a task. 
            \item Past resource consumption of the MC agent. \end{itemize}
            \item Model of the world, which contains information about:
            \begin{itemize}
\item Tasks, e.g., their input, goal, skills needed to solve it, etc.
\item Decision environment. 
\item Actions available to the solvers in each state and for each task.
\end{itemize}
\item Model of others, which maintains knowledge and beliefs over other agents that may act in the same decision environment. 
          \end{itemize}
        \item S1 solver(s):
\begin{itemize}
    \item Proposed decision 
    for a task.
\item 
Confidence in the proposed decision. 
\end{itemize}
\end{enumerate}

The first meta-cognitive phase (MC1) 
activates automatically as a new task arrives and a solution for the problem is provided by an S1 solver. 
MC1 decides between accepting the solution proposed by the S1 solver or activating the second meta-cognitive phase (MC2).
MC1 makes sure that there are enough resources, specifically time and memory, for completing
both MC1 and MC2. If not, MC1 adopts the S1 solver's proposed solution.
MC1 also compares the confidence provided by the S1 solver with the expected reward for that task: if the confidence is high enough compared to the expected reward, MC1 adopts the S1 solver’s solution. Otherwise, it activates the next assessment phase (MC2) to make a more careful decision.
The rationale for this phase of the decision process is that we envision that often the system will adopt the solution proposed by the S1 solver, because it is good enough given the expected reward for solving the task, or because there are not enough resources to invoke more complex reasoning. 

Contrarily to MC1, MC2 decides between accepting the solution proposed by the S1 solver or activating an S2 solver to solve the task. 
To do this, MC2 evaluates the expected reward of using the S2 solver in the current state to solve the given task, using information contained
in the model of self about past actions taken by this or other solvers to solve the same task, and the expected cost of running this solver.
MC2 then compares the expected reward for the S2 solver with the expected reward of the action proposed by the S1 solver: if the expected additional reward of running the S2 solver, as compared to using the S1 solution, is greater than the expected cost of running the S2 solver, then MC2 activates the S2 solver. Otherwise, it adopts the S1 solver's solution. 

To evaluate the expected reward of an action (here used to evaluate the S1 solver's proposed action), MC2 retrieves from the model of self the expected immediate and future reward for the action in the current state 
(approximating the forward analysis to avoid a too costly computation),
and combines this information with the confidence the S1 solver has in the action.
The rationale for the behavior of MC2 is based on the design decision to avoid costly reasoning processes unless the additional cost is compensated by an even greater additional expected value for the solution that the S2 solver will identify for this task, analogous to what happens in humans \cite{Shenhav2013}.

For the sake of simplicity and lack of space, in this paper we did not consider the possibility of having several S1 and/or S2 solvers for the same task, which would bring additional steps for both MC1 and MC2 to identify the most suitable solver.
We also did not consider issues related to problem similarity, and we assumed we have
enough past experience 
about exactly the same problem 
to be able to evaluate the expected reward and utility of solvers and actions.
Finally, we did not discuss how the expected value of an action and the associated confidence are combined: currently we are working with a risk-adverse approach where these two quantities are multiplied, but we also plan to explore other approaches.


\section{Instances of the SOFAI Architecture}

As mentioned, we believe that many real-life settings present sequential decision problems. Available solvers could be able to choose one single decision of a whole sequence, or they may be able to build a whole decision sequence (or everything in between).
In the first scenario, the MC agent is used to select the appropriate solver for each decision. In the second setting, the MC agent is used only once, so its influence on the performance of the system and the quality of its decisions could be more limited. However, usually the second setting occurs when we have solvers with a deep knowledge of the decision domain, and can adapt their capabilities to the available resources and expected reward for optimality and correctness of the decision sequences.
We envision that these parameters and domain knowledge can be exploited by the meta-cognitive agent both in the MC1 and the MC2 phase.

We are currently working on implementing two main instances of the SOFAI 
architecture, that cover both extremes of this spectrum.
In the first instance, we consider a decision environment in the form of a constrained grid (where constraints are over states, moves, and state features), where solvers are only able to decide on the next move but entire trajectories (from an initial state to a goal state) need to be built to solve the given task \cite{glazier2021making}.
In the second instance, we consider epistemic planning problems \cite{Fagin2003} where solvers are indeed planners, so they are able to build an entire plan and not just a single step in the plan. In this case, the meta-cognitive module includes also steps regarding problem and solver simplifications, as well as problem similarity assessment.

\bibliographystyle{abbrv}
\bibliography{biblio}

\end{document}